\documentclass[fullpaper,cameraready]{nldl}

\renewcommand{\DOI}{10.7557/18.6805}

\usepackage[utf8]{inputenc}
\usepackage{url}
\usepackage{graphicx}
\usepackage{authblk}
\usepackage{algorithm}
\usepackage{listings}
\usepackage{color,soul}
\usepackage[noend]{algpseudocode}
\usepackage[square,numbers]{natbib}

\usepackage[pagebackref]{hyperref}

\title{Learning to solve arithmetic problems with a virtual abacus}
\author[1$\dagger$]{Flavio Petruzzellis}
\author[1$\dagger$]{Ling Xuan Chen}
\author[1]{Alberto Testolin\thanks{Corresponding Author: alberto.testolin@unipd.it}}
\affil[1]{University of Padova, Italy}
\affil[$\dagger$]{Equal contribution}

\date{\vspace{-5ex}}

\begin{document}
\nldlmaketitle

\begin{abstract}
Acquiring mathematical skills is considered a key challenge for modern Artificial Intelligence systems. Inspired by the way humans discover numerical knowledge, here we introduce a deep reinforcement learning framework that allows to simulate how cognitive agents could gradually learn to solve arithmetic problems by interacting with a virtual abacus. The proposed model successfully learn to perform multi-digit additions and subtractions, achieving an error rate below 1\% even when operands are much longer than those observed during training. We also compare the performance of learning agents receiving a different amount of explicit supervision, and we analyze the most common error patterns to better understand the limitations and biases resulting from our design choices.
\end{abstract}

\section{Introduction}
Deep learning systems excel in a variety of domains, but struggle to learn cognitive tasks that require the manipulation of symbolic knowledge \cite{lake_ullman_tenenbaum_gershman_2017}. This limitation is particularly evident in the field of mathematical cognition \cite{10.3389/fnhum.2020.00100}, which requires to grasp abstract relationships and deploy sophisticated reasoning procedures. Indeed, even state-of-the-art language models fall short in mathematical tasks that require strong generalization capabilities \cite{DBLP:conf/iclr/SaxtonGHK19} (though very recent work has shown that performance significantly improves following fine-tuning on large-scale mathematical datasets \cite{lewkowycz2022solving}).

One possibility to tackle this challenge is to endow deep architectures with \textit{ad-hoc} primitives specifically designed to manipulate arithmetic concepts \cite{NEURIPS2018_0e64a7b0, Madsen2020Neural}. Nevertheless, alternative approaches suggest that symbolic numerical competence could emerge from domain-general learning mechanisms \cite{DBLP:journals/corr/KaiserS15, cognolato2022transformers}. Recent work has also tried to deploy deep reinforcement learning (RL) to solve math word problems \cite{Wang_Zhang_Gao_Song_Guo_Shen_2018}. Notably, deep RL architectures that incorporate copy and alignment mechanisms seem to discover more sophisticated problem solving procedures \cite{huang-etal-2018-neural}, and could even learn automatic theorem proving when combined with Monte-Carlo tree search algorithms \cite{NEURIPS2018_55acf853}.

In this work we explore whether model-free deep RL agents could learn to solve simple arithmetic tasks (expressions involving sum and subtraction) by exploiting an external representational tool, which can be functionally conceived as a \textit{virtual abacus}.
Differently from the above-mentioned approaches, our goal is to take advantage of the way humans solve the task by simulating the interaction of the RL agent with an abacus, which can be partially guided through supervised learning mechanisms. This allows teaching the agent existing algorithms for the solution of arithmetic problems, rather than forcing it to discover possible solution strategies only by trial-and-error.
Our work is similar in spirit to recent scientific endeavors directed towards the design of learning systems that are capable of the kind of systematic generalization that is required to execute algorithmic procedures \citep{Veli2020Neural, NEURIPS2021_3501672e}.
The main challenge of the problem is to discover an algorithm that can be used to solve arithmetic problems, and thus being able to generalize to never-seen instances of the same class of tasks.
In a deep RL framework, this even more difficult since rewards could become very sparse due to the length of the solution procedures.

In particular, we are interested in addressing these two key questions: Is it possible to learn to solve mathematical problems that require long-term planning by only relying on model-free RL? If not, what is the minimal amount of guidance (in the form of learning biases and/or explicit supervision) that is necessary to successfully solve such problems?

We find that our agent is able to solve the sum and subtraction problems with a considerable capacity of out-of-distribution (OOD) generalization over the length of the operands. At the same time, our simulations shed light on difficulties in solving mathematical problems with model-free RL, especially when learning requires to plan in the very distant future or to discover solution strategies in which simple steps should be combined to solve arbitrarily complex problems. By systematically analysing the errors made by the agent in the OOD generalization regime, we also provide some intuitions about the functioning and limitations of the proposed learning framework.

\section{Methods}
In this section we describe the task to be solved, the design of the environment with which the agent interacts, the agent architecture, and the training procedure.

\subsection{Task}
The task of the agent is to compute arithmetic operations between integers, which are provided as a sequence of input symbols that can be either an operation (plus or minus) or an operand. Operands are represented as sequences of digits: during training, the length of the operand is sampled uniformly in $\{1, 2, 3, 4, 5, 6\}$, and each digit is sampled uniformly in $\{0, 1, 2, 3, 4\}$, except for the first digit which cannot be 0. The first operand of any operation is always the one represented in the current state of the abacus, whose initial configuration represents the value 0.

\begin{figure}
    \centering
    \includegraphics[width=0.7\linewidth]{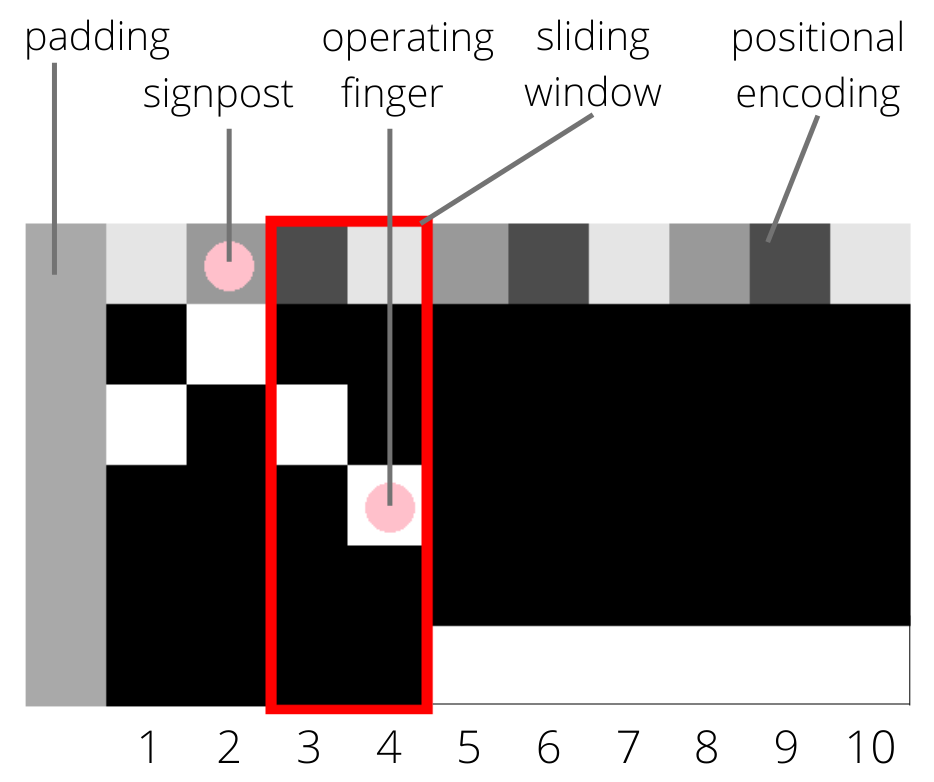}
    \caption{The environment consists of a simulated abacus with 10 columns, a padding on the left and a periodic positional encoding on top. The agent observes the abacus through a sliding window composed of two columns, interacts with it using an `operating finger' and uses a `signpost' to mark the column where the current operation is going on.}
    \label{fig:env}
\end{figure}

\subsection{Environment}
The learning environment is conceived as a simulated abacus, featuring 10 columns with 5 positions for each column: that is, we represent numbers in base 5 (see Fig. \ref{fig:env}).
On the top edge of the abacus there is an additional row representing a periodic positional encoding, which allows to associate each column with a value in the sequence [0.25, 0.5, 0.75, 0.25 ...]. Such additional input allows to effectively encode the position of the fingers in the abacus (as explained below), thus improving learning speed and generalization.

The agent can manipulate the abacus using an `operating finger', which serves as a pointer positioned over the abacus at a given location at every time step.
Furthermore, the agent can use an indicator (dubbed the `signpost') that can be used to signal the column where the current operation is occurring, and where the operation must resume once a carry operation is over.

The agent partially observes the abacus through a sliding window composed by two columns: the one where the operating finger is, and the one to its left. If the signpost is in the sliding window, the agent can see it in superposition to the positional encoding.
In case the operating finger is on the first column, the agent observes a padding instead of the column to the left of the operating finger.

The agent can interact with the environment performing one of the following actions: a movement of the operating finger in the four directions, a movement of the signpost along two directions (left, right), a slide of the beads in the column where the operating finger is currently positioned (dubbed the `move and slide' action) and an action to signal that it has finished processing the current digit (dubbed the `submit' action).

When the agent acts, the environment changes its state and gives as output the reward and a flag indicating whether the current episode is over.
At each time step, the agent receives as input either an operation symbol or the next digit that should be processed, both represented as one-hot vectors.
The operation is also signalled using a flag throughout the duration of the operation.
The episode ends if the abacus reaches its maximum representational capacity, or if terminated early by the environment (see section \ref{subsec:task-reward}).

\subsection{Agent's architecture}
We simultaneously train an actor and a critic network.
Both are memory-less feed-forward networks and thus receive as input a stack of the last 3 observations, processed by a feature extractor implemented as a 3D convolutional network with three layers of 128, 255 and 512 channels, respectively.
We use kernels of size 3 in the first layer and size 2 in the last two, all with padding of 1.
The output of the feature extractor is then concatenated with the one-hot encoded symbol received from the environment; the critic also receives as input the previous action.
Both networks then process this input via 5 feed-forward layers with 2048, 1024, 512, 256 and 128 neurons, respectively\footnote{We have chosen all hyper-parameters for the feature extractor and the feed-forward networks starting from small architectures and increasing their size until we achieved a satisfactory performance.}.
Finally, the actor network returns a probability distribution over the actions, while the critic returns an estimate of the value of each action.

\subsection{Reward function}
\label{subsec:task-reward}
We define specific algorithms to solve the sum and subtraction problems using the virtual abacus, and use such ideal solutions to provide supervision to the learning agent (pseudo-code is provided in Algorithm \ref{alg:add})\footnote{We do not report the subtraction algorithm since it only differs in lines 4, 7, 13 and 14 where we implement the actual operation and check if a carry is necessary.}. In order to encourage the agent to learn such algorithms, we designed a modular reward function that makes it possible to provide an increasing amount of supervision through the following feedbacks:

\begin{itemize}
    \item A penalty of -0.05 for each action performed, to discourage long sequences of actions.
    \item A reward of 0.10 whenever a movement action (left, right, down, up) moves the operating finger into a position that is closer to the target algorithmic solution. A penalty of -0.10 is given if the opposite happens. \item A reward of 1 if the signpost correctly moves to the next position when a partial sum is done, or when the agent correctly resets the signpost.
    \item A reward of 1 for correct move and slide.
    \item A reward of 1 when the agent chooses the submit action, and the abacus configuration represents the correct partial result.
    \item If the agent does any of the previous three actions in the wrong way, the episode is terminated and the agent is given a penalty of -1.
\end{itemize}

\begin{algorithm}
\scriptsize
\caption{Addition algorithm. 
$c_r$ and $c_l$ are the two columns (left and right) visible to the agent in the sliding window.
$S(c)$ is a function that returns the symbol encoded in a column.}\label{alg:add}
    \begin{algorithmic}[1]
    \While{not done}
        \State Read symbol $s$ from environment
        \If{$s$ is digit}
            \State Write $(S(c_r)+s) \% 5$ on $c_r$
            \If{signpost in $c_l$}
                \State Move signpost right
            \ElsIf{$S(c_r)+s \geq 5$}
                \State carry $\gets True$
            \Else
                \State carry $\gets False$
            \EndIf
            \While{carry $= True$}
                \State Move operating finger right
                \State Write $(S(c_r)+s) \% 5$ on $c_r$
                \If{$S(c_r)+1 \geq 5$}
                    \State carry $\gets True$
                \Else
                    \State carry $\gets False$
                \EndIf
            \EndWhile
            \While{signpost not in $c_l$}
                \State Move operating finger left
            \EndWhile
        \Else \algorithmiccomment{Reset the abacus}
            \While{$c_l$ is not padding}
                \State Move operating finger left
            \EndWhile
            \While{signpost not in $c_r$}
                \State Move signpost left
            \EndWhile
        \EndIf
    \EndWhile
    \end{algorithmic}
\end{algorithm}

We determine the maximum length of an episode dynamically, in order to avoid long loops of meaningless actions. The agent is given 32 timesteps for each correct move and slide action - i.e., the maximal theoretical distance between any two possible move and slide actions in a solution trajectory according to the reference algorithm.
This mechanism grants the agent enough timesteps to complete the operations that have long trajectories (i.e. resetting the abacus at the end of an operation or computing a long carry), while also allowing the agent to explore the functioning of the virtual abacus during training.

\subsection{Model training}
\label{sec:train-test}
We use the Proximal Policy Optimization (PPO) learning algorithm with a linearly-decaying sinusoidal learning rate, clipped surrogate objective funcion \cite{https://doi.org/10.48550/arxiv.1707.06347}, frame stacking \cite{https://doi.org/10.48550/arxiv.1312.5602} and masking \cite{https://doi.org/10.48550/arxiv.1708.04782, Huang_Ontanon_2022}, as implemented in the Python framework Stable Baselines3 \cite{JMLR:v22:20-1364}.
We have also tried to apply the DQN learning algorithm \cite{https://doi.org/10.48550/arxiv.1312.5602} with frame stacking and the same learning rate decay scheme we used with PPO. However, we observed an extremely slow speed of convergence in all reward settings, and especially with the ones having sparse rewards.

Past information is provided by feeding the last 3 environment states stacked into a 3D tensor.
We mask illegal actions, e.g. moving left on the left edge of the board, or actions that do not produce any effect on the environment, such as using move and slide when the abacus is already in the target configuration.
We implemented an early stopping criterion, whereby learning is interrupted if the KL divergence between the old policy and the new one is greater than a threshold $\alpha=0.2$ that was empirically chosen.

\section{Results}
In this section we describe the simulation results and we provide some insights in the way the agent works by analysing its failures when probed to solve problems involving integers longer than the ones seen during training.
\footnote{We make the code that was used to run the experiments publicly available at this GitHub repository: \hyperlink{https://github.com/ChenEmmaL/imitation_abacus}{\texttt{https://github.com/ChenEmmaL/imitation\_abacus}}.}

\subsection{Solving arithmetic tasks}
We designed the simulations with the aim of studying the level of supervision that is necessary to successfully learn the arithmetic task.
To this aim, we trained the agent with varying amounts of reward: in the simplest case, we included all components of the modular reward function.
We then removed the components of the reward related to the movement of the operating finger (OF) and those related to the movement of the signpost (SP).
The first component provides very frequent but not strictly necessary supervision, as the agent can discover the correct movement of the operating finger by exploration.
The second component provides important information to the agent, in that the correct movement of the signpost is necessary to solve operations that require (possibly very long) carries.

\begin{figure}
    \centering
    \includegraphics[width=\linewidth]{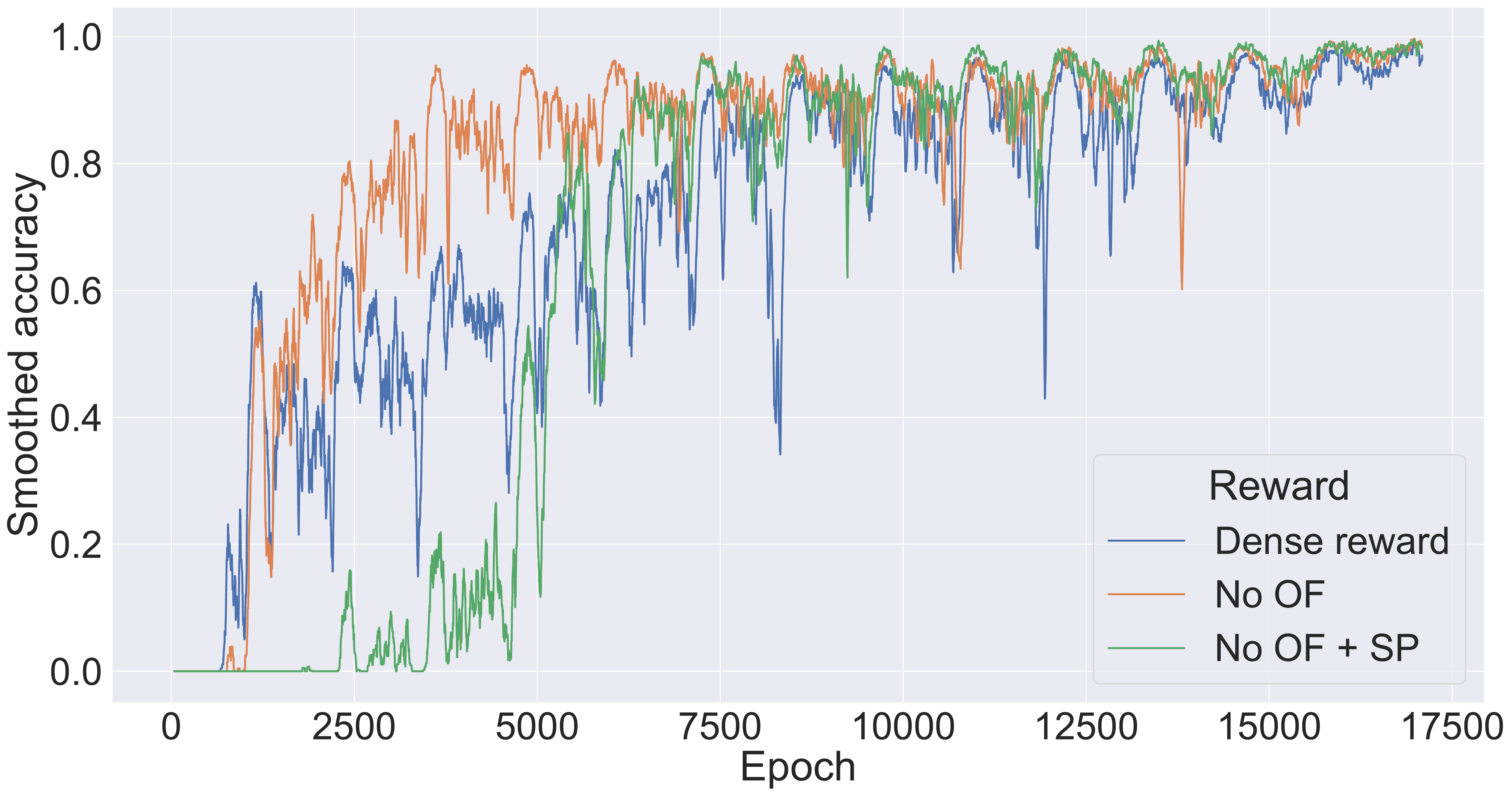}
    \caption{Learning performance of models trained with varying amounts of supervision (oscillations are due to the sinusoidal learning rate). We measure accuracy as the fraction of operations correctly computed in one epoch.}
    \label{fig:acc}
\end{figure}

As shown in Fig. \ref{fig:acc}, the agent is able to solve the task in all cases, reaching an almost perfect accuracy. Surprisingly, reducing the amount of supervision and letting the agent discover the most effective way to use the operating finger leads to a \textit{faster} training and also higher performance in terms of number of consecutive operations successfully computed (see Table \ref{tab:max-op}).
Further reducing the level of supervision causes an initial learning slowdown but still allows to reach a very high accuracy later on, although the final performance in terms of number of consecutive operations is lower compared to the intermediate level of supervision.

Next, we have removed the reward for correct move and slide actions, observing that learning becomes so slow that the performance at every epoch is barely improving \footnote{We did not remove the penalty for long trajectories as it shapes the behavior of the agent only indirectly.
We also did not remove the component of the reward that signals if a partial operation was successfully completed, as it provides the agent with the most important signal about the completion of the task.}.
Therefore, we find that in order to learn the algorithm in a reasonable time, the agent needs to receive the following essential feedback: the reward for correct move and slides, the reward for correct partial answers given using the `submit' action, the penalty for long trajectories, and the penalty (including episode termination) in case of wrong move and slides or partial answers.

\begin{figure}
    \centering
    \includegraphics[width=\linewidth]{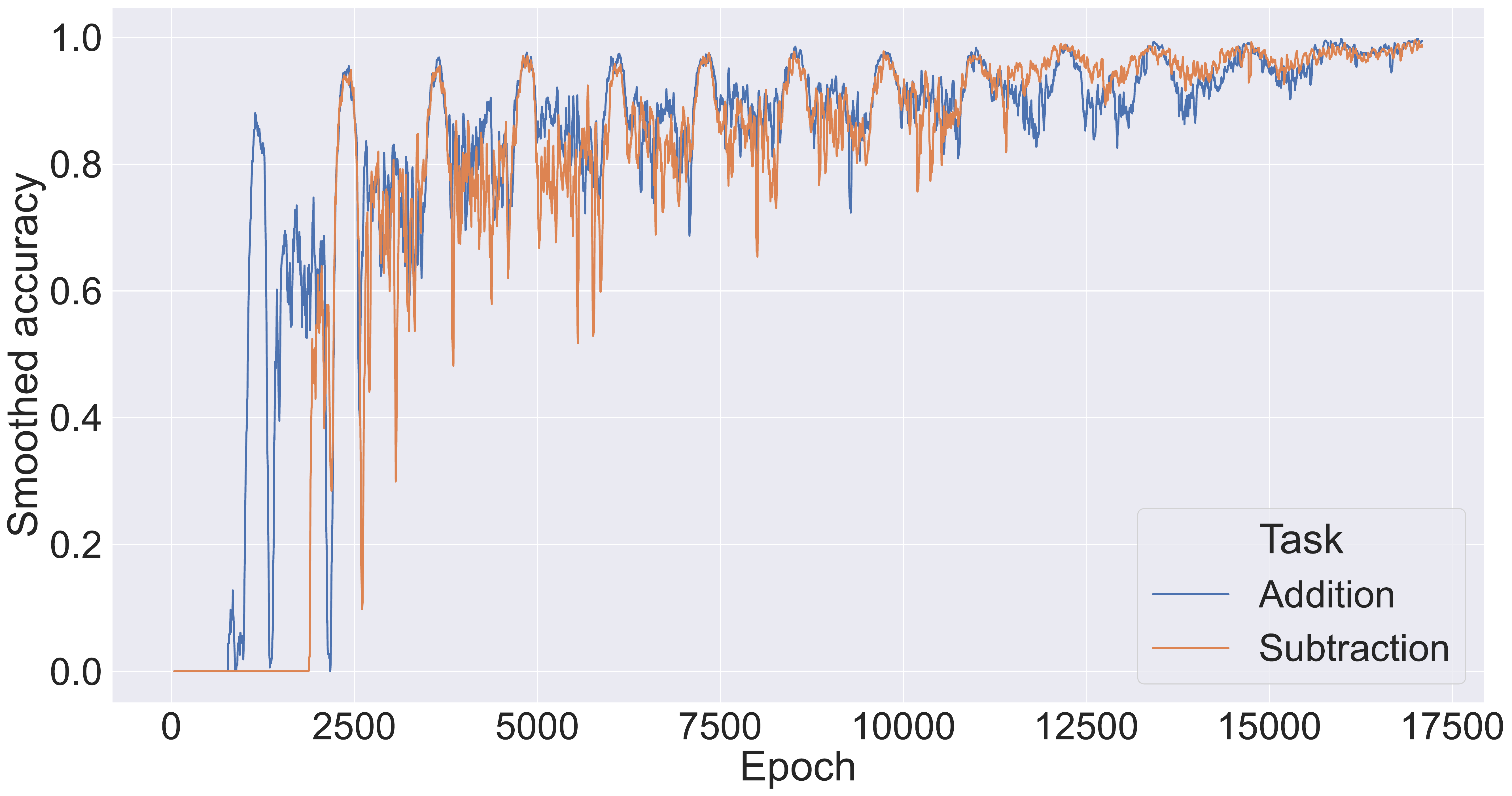}
    \caption{Learning performance of the models trained on sum only or subtraction only with the dense reward function.}
    \label{fig:acc-sep}
\end{figure}

We also trained the agent on each arithmetical task separately, using the highest level of supervision. As expected, learning a single task is simpler, both in terms of learning speed (Fig. \ref{fig:acc-sep}) and number of successful consecutive operations (Table \ref{tab:max-op}).

\subsection{Generalizing to longer operands}
Learning to solve arithmetic operations with operands ranging beyond the intervals encountered during training is one of the main challenges for deep learning models \cite{cognolato2022transformers}.
We thus investigated the capability of our best performing agent, namely the one trained on both tasks with an intermediate level of supervision\footnote{Note that in this case we extend the abacus to 20 columns in order to represent the extended range of operands.} by sampling the two test operands in the intervals $[5^{x-1}, 5^x)$, where $x \in \{1, 2, 4, 8, 16\}$.

\begin{figure}
    \vspace{0.1cm}
    \centering
    \includegraphics[width=0.9\linewidth]{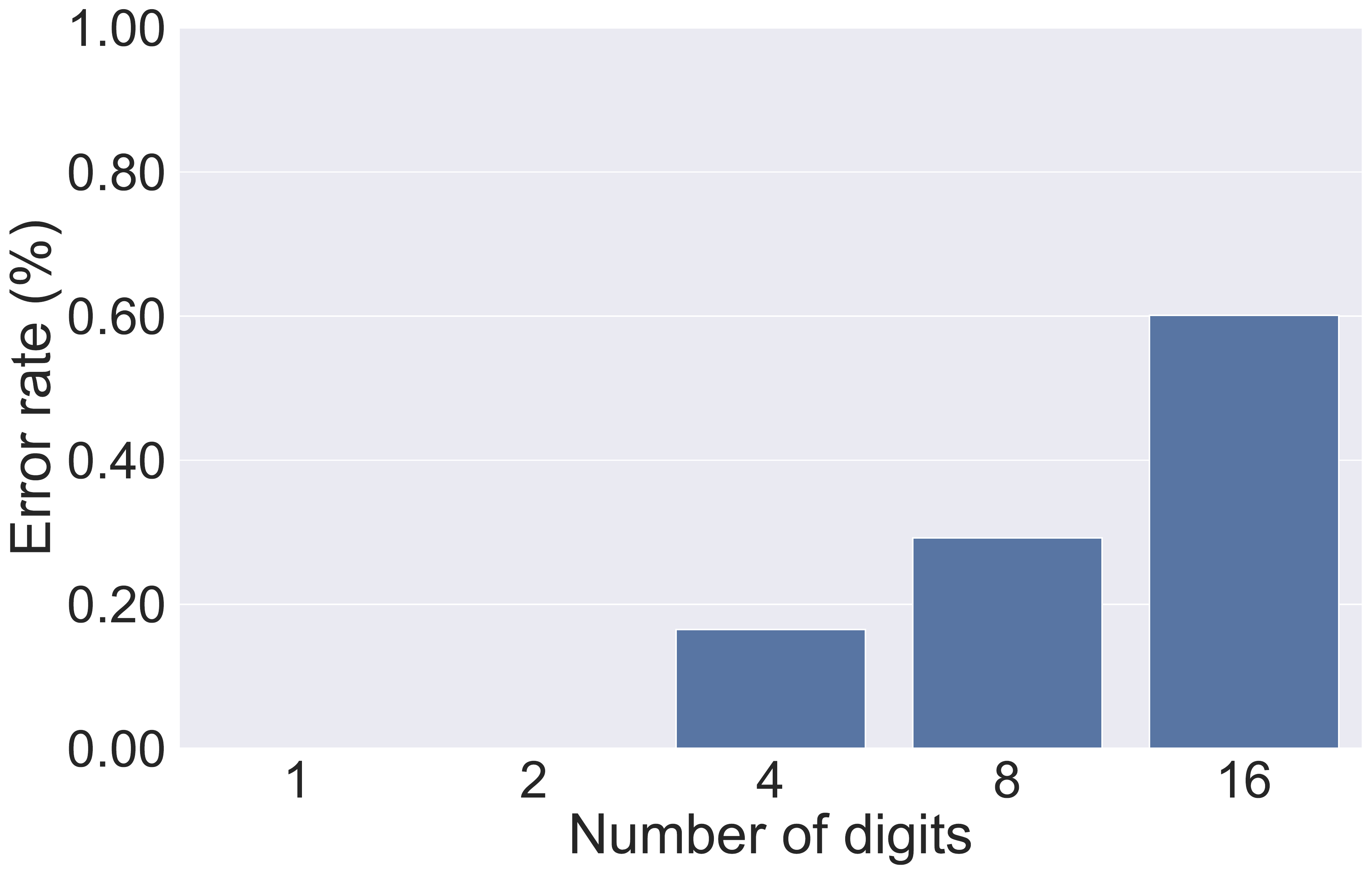}
    \caption{Error rate of the model on operations involving longer operands than those seen during training, which contained at most 6 digits.}
    \label{fig:ood-digits}
\end{figure}

For each interval, we sampled 100000 operands and counted the number of mistakes made by the agent. Although the trained agent does not exhibit perfect OOD generalization capabilities and its error rate grows with the number of digits involved in the operations (see Fig. \ref{fig:ood-digits}), it can still solve sums and differences involving operands in intervals unseen during training with an arguably low error rate (e.g., when operands involved more than two times the amount of digits observed during training, the error rate was still below 1\%).

\begin{table}
    \centering
    \begin{tabular}{|l|c|}
        \hline
        \textbf{Model} & \textbf{N. op.} \\
        \hline
        Dense reward & 515 \\
        No OF & 1145 \\
        No OF + SP & 773 \\
        \hline\hline
        Addition & 907 \\
        Subtraction & 675 \\
        \hline
    \end{tabular}
    \caption{Maximum number of operations successfully completed during training in the different training scenarios.}
    \label{tab:max-op}
\end{table}

\subsection{Analysis of errors}
We analyzed the pattern of errors made in the most extreme OOD regime, that is, when operands were sampled in the interval $[5^{15}, 5^{16})$, by collecting statistics over 100000 simulations. The agent commits 601 errors, which is coherent with the error rate reported in Fig. \ref{fig:ood-digits}. We recorded the errors and manually compiled a list of 4 different error classes: errors in the movement of the signpost, errors during a carry operation, and errors occurring during `simple' addition or subtraction operations. The latter kind of errors might in fact include operations that require a carry; we only selected errors that did not happen when performing the carry operation (e.g. removing one unit from the column to the right and fill the current column) but instead occurred when the carry was completed and the agent needed to compute a sum or subtraction.

\begin{table}
    \centering
    \begin{tabular}{|l|c|}
        \hline
        \textbf{Error class} & \textbf{\%} \\      
        \hline
        Simple operation & 32.9\\
        Signpost right & 27.7 \\
        Carry & 27.3 \\
        Signpost left & 12.1 \\
        \hline
    \end{tabular}
    \caption{Relative frequency of error types. A simple operation is a sum or subtraction between two digits. The Carry and Signpost right classes occur when the agent must compute a carry operation.}
    \label{tab:err-cat}
\end{table}

As shown in Table \ref{tab:err-cat} the most frequent kind of error is the one involving a simple sum or subtraction operation.
The second and third most common classes of errors are wrong carries or movements of the signpost rightwards: notice that such a movement is required precisely when the agent must perform a carry operation.
These results reflect the fact that the carry operation is the most complicated step in multi-digit sum or subtraction.
Lastly, the least frequent kind of error is the one involving the movement of the signpost to the left, indicating that the agent has learned to reset the abacus to the initial position almost perfectly.

\begin{figure}
    \center
    \includegraphics[width=\linewidth]{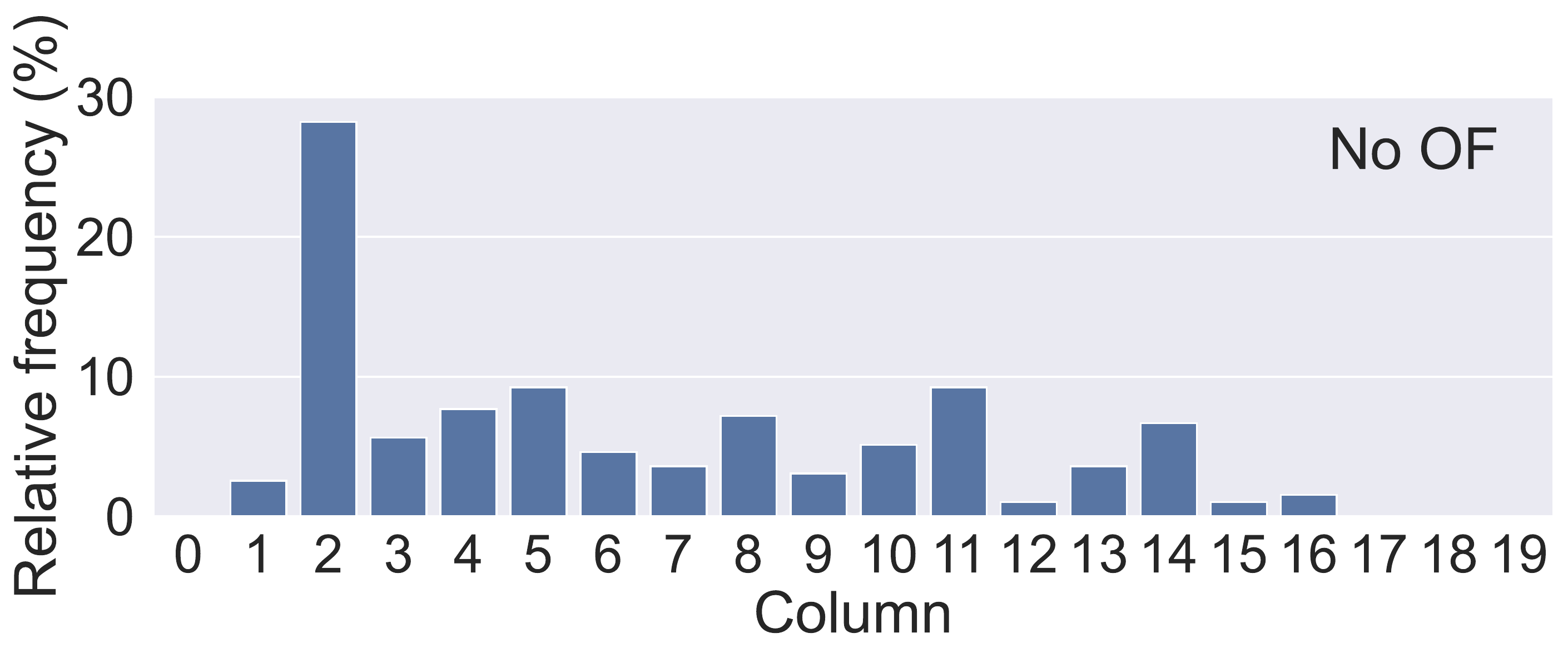}
    \caption{Relative frequency of errors by column of the abacus for the best model. Most errors happen on the second column, and a periodic pattern emerges from the third column onward.}
    \label{fig:err-col}
\end{figure}

Since our system includes a sliding window mechanism that limits the capacity of the agent to observe the abacus, we investigated whether errors are equally distributed on all columns. 
The histogram in Fig. \ref{fig:err-col} representing the frequency of errors by column shows that most errors occur on the second column, which is the first one observed outside the initial position of the sliding window. 
This is consistent with our previous observation that many errors occur during a carry operation, which requires to move the sliding window to the right.
We can also observe a periodic pattern of errors from the third column onward, most likely due to the specific choice used in the positional encoding.
This reveals that, although this element of the environment contributes to the capacity of the agent to generalize to unseen ranges of operands, it also introduces a regularity in the errors which depends on the column where an operation must be computed.

Finally, in Table \ref{tab:err-example} we report a few significant examples of errors made by our system in the OOD generalization regimen.
Consistently with the previous analysis, we can see that for both sums and subtractions the agent can make a mistake in the early positions (first columns) as well as in the last ones.
Also, it is evident that an error in a carry or regrouping can propagate to the following columns.
However, it can also happen that the system is resilient to such mistakes, and thus still compute the rest of the operation correctly: this is a desirable property, since it avoids propagating errors and thus keeps the absolute value of the error relatively low.

\begin{table}
    \vspace{0.1cm}
    \centering
    \begin{tabular}{l|r|r}
        \textbf{S} & 2204010440402424 & 3014100322010344 \\ 
        \textbf{I} & +3422111404102300 & +1342122200324413 \\
        \textbf{O} & 11131122400010\hl{0}24 & \hl{30141003}22340312 \\
        \textbf{T} & 11131122400010224 & 4411223022340312 \\
        \hline
        \hline
        \textbf{S} & 4110012024223441 & 3342443241400324 \\
        \textbf{I} & -1300014214322224 & -3231303122242113 \\
        \textbf{O} & \hl{4041}442304401212 & 111\hl{0}40114103211 \\
        \textbf{T} & 2304442304401212 & 111140114103211 \\
    \end{tabular}
    \caption{Examples of errors. We report the state of the abacus (\textbf{S}), operation input to the system to be executed (\textbf{I}), output produced (\textbf{O}) and true output (\textbf{T}). Operands are in base 5 to facilitate the interpretation of mistakes in carries and regroupings.}
    \label{tab:err-example}
\end{table}

\section{Discussion}
In this work we introduced a framework that can be used to study how an agent can learn to solve arithmetic operations by exploiting deep reinforcement learning and by interacting with external representations that incorporate the working principles of an abacus.
Our framework is inspired by the way humans interact with external representational tools to solve mathematical problems, connecting to the more general trend of exploring tool use in deep reinforcement learning environments \cite{Baker2020Emergent, 10.1007/978-3-030-19642-4_23, sabathiel2022self}.
Differently from similar problems in the deep RL literature, such as learning to play combinatorial games, the problem we propose is characterized by an algorithmic nature, in that the goal of the agent is learning an exact solution algorithm to arithmetic problems, rather than a strategy to win a game.

Our simulations suggest that in order to learn to solve arithmetic problems with model-free reinforcement learning, a memory-less agent needs to receive a certain amount of explicit supervision to overcome its inability to plan in the long-term.
Notably, the agent we present is able to solve problems involving operands that are well outside the training range, which can be considered the main challenge of the class of arithmetic problems we propose.
The agent is able to do so thanks to specific learning biases: a sliding window, a positional encoding, and the representation format of the operands on the abacus.

It might be possible let the agent fully observe the virtual abacus and learn which part is relevant in any given moment; however, by adopting a relative view on the learning environment through the sliding window the agent can effortlessly generalize the strategy learned on a limited interval of operands to ones that are more than two times longer. At the same time, it turned out that including elements contributing to generalization capability, such as the periodic positional encoding, also introduced unwanted regularities in the errors committed by the system.

An exciting venue for future research would be to explore the possibility to endow the agent with some form of memory (e.g., by exploiting recurrent architectures), which would allow to plan in the distant future and thus learn to solve the problem with even less supervision.
Furthermore, it would be interesting to exploit model-based and hierarchical reinforcement learning approaches to endow the agent with native capability to internally simulate the functioning of the external tool and learn to compose simple solution steps into more complex strategies.


\end{document}